\documentclass[letterpaper]{article} 
\usepackage{aaai25}  
\usepackage{times}  
\usepackage{helvet}  
\usepackage{courier}  
\usepackage[hyphens]{url}  
\usepackage{graphicx} 
\usepackage{natbib}  
\usepackage{caption} 
\usepackage{amsmath}
\usepackage{amssymb}
\usepackage{booktabs}
\usepackage{placeins}
\usepackage{array}
\usepackage{newfloat}
\usepackage{listings}
\usepackage{subcaption}
\usepackage[percent]{overpic}
\usepackage{csquotes}
\usepackage{float}
\usepackage{enumitem}
\usepackage{hyperref}
\usepackage[linesnumbered,ruled,vlined]{algorithm2e} 

\usepackage{newfloat}
\usepackage{listings}
\DeclareCaptionStyle{ruled}{labelfont=normalfont,labelsep=colon,strut=off} 
\floatstyle{ruled}
\newfloat{listing}{tb}{lst}{}
\floatname{listing}{Listing}
\title{Generating Traffic Scenarios via In-Context Learning \\ to Learn Better Motion Planner}
\author{
    Aizierjiang Aiersilan
}
\affiliations{
    University of Macau\\
    ezharjan@outlook.com
}

\begin{document}
\maketitle

\begin{abstract}
Motion planning is a crucial component in autonomous driving. State-of-the-art motion planners are trained on meticulously curated datasets, which are not only expensive to annotate but also insufficient in capturing rarely seen critical scenarios. Failing to account for such scenarios poses a significant risk to motion planners and may lead to incidents during testing. An intuitive solution is to manually compose such scenarios by programming and executing a simulator (e.g., CARLA). However, this approach incurs substantial human costs. Motivated by this, we propose an inexpensive method for generating diverse critical traffic scenarios to train more robust motion planners. First, we represent traffic scenarios as scripts, which are then used by the simulator to generate traffic scenarios. Next, we develop a method that accepts user-specified text descriptions, which a Large Language Model translates into scripts using in-context learning.
The output scripts are sent to the simulator that produces the corresponding traffic scenarios. As our method can generate abundant safety-critical traffic scenarios, we use them as synthetic training data for motion planners. To demonstrate the value of generated scenarios, we train existing motion planners on our synthetic data, real-world datasets, and a combination of both.
Our experiments show that motion planners trained with our data significantly outperform those trained solely on real-world data, showing the usefulness of our synthetic data and the effectiveness of our data generation method.
\end{abstract}

\begin{links}
\link{Code}{https://ezharjan.github.io/AutoSceneGen}
\end{links}

\label{sec:formatting}

\section{Introduction}
\label{sec:intro}

Efficiently evaluating autonomous vehicles (AVs) under diverse real-world challenges on a limited budget is crucial for ensuring their safety and sustaining long-term growth in the AV industry. While existing works on vehicle or pedestrian trajectory prediction, such as TrafficPredict \cite{ma2019trafficpredict}, Pishgu \cite{alinezhad2023pishgu}, TraPHic \cite{chandra2019traphic}, and MSRL \cite{wu2023multi}, rely on real-world datasets like ApolloScapes \cite{huang2018apolloscape}, NGSIM \cite{kovvali2007video}, UCY \cite{lerner2007crowds}, and ETH \cite{pellegrini2009you}, which may not fully capture the complexities of modern trajectory prediction scenarios involving both AVs and non-AVs, our custom virtual dataset, generated using our framework \enquote{AutoSceneGen}, produces improved prediction results. Most importantly, generating and synthesizing data using our prompt-based, configurable, and AI-driven AutoSceneGen framework is more cost-effective and efficient than collecting real-world datasets such as Argoverse2 \cite{wilson2023argoverse}, Waymo \cite{ettinger2021large}, Round \cite{krajewski2020round}, Ind \cite{bock2020ind}, nuScenes \cite{caesar2020nuscenes}, Argoverse \cite{chang2019argoverse}, Highd \cite{krajewski2018highd}, and ApolloScapes. Our framework automatically generates datasets with real-world features and provides easy control over dataset heterogeneity through scenario descriptions, especially for safety-critical scenarios.

{\bf Status quo.} 
Despite the challenges and inefficiencies in the testing phase of AVs, significant progress has been made in optimizing the testing of individual modules, such as motion planners. However, most efforts have focused on evaluating trajectory prediction models in simulation environments \cite{shet2023path, li2023safe}. While simulations reduce testing costs related to budget and safety, creating realistic and logically specific scenarios often requires substantial human effort, though still less than testing AVs on real roads. Additionally, safety evaluation is typically performed module-by-module, demanding considerable resources to create safety-critical scenarios within simulators.
Scenario-based evaluation methods have largely focused on coverage, unsafe scenarios, and indicator-estimation tests \cite{jesenski2019generation, amersbach2019defining}. Recent advancements in scenario generation prioritize data-driven methods \cite{sun2024drivescenegen}, with generative models trained on domain-specific data for traffic scenario generation \cite{tan2023language}. However, these methods struggle with low-resource scenarios, such as corner cases and accidents, due to limited data availability.
There is a clear need for a universal, general, and budget-conscious framework that enhances traffic scenario heterogeneity automatically through scenario descriptions. This would expedite the simulation and testing process. While advanced AI tools, like large language models (LLMs), show promise, they often require extensive post-processing to correct inaccuracies or generate overly generic results. Our framework addresses these issues by automating complex scenario generation in a cost-effective manner, overcoming the limitations of low-resource scenarios and optimizing the use of simulator functionalities to ensure scenarios are diverse, realistic, and reflective of challenging real-world situations, as highlighted by \citet{bogdoll2021desccorner}.

{\bf Contributions.} Scenario generation has traditionally been a manual process requiring significant human effort. However, with the advancement of LLMs, there is now an opportunity to leverage AI to efficiently generate specific traffic scenarios. Building on prior research, the main contributions of this study are as follows:

\begin{itemize}[left=0pt]
    \item A universal, general, and cost-effective framework, \enquote{AutoSceneGen}, is proposed to automatically enhance the heterogeneity of traffic scenarios through scenario descriptions, thereby accelerating the simulation and testing process. 
    \item AutoSceneGen leverages in-context learning (ICL) of LLMs, eliminating the need for training or fine-tuning generative models for scenario generation tasks.
    \item The scenarios generated by AutoSceneGen were demonstrated to produce better datasets, leading to improved training results for motion planners.
    \item AutoSceneGen automatically categorizes the generated scenarios based on their descriptions, eliminating the need for downstream annotation and facilitating the training of motion planners in open-world scenarios.
    \item AutoSceneGen is a modular framework with dynamic components, allowing easy substitution of its generative model and simulation engine used for scenario generation and data collection.
\end{itemize}

\section{Related Work}

\subsection{Scenario Generation and Gaps}
Existing works, such as LCTGen \cite{tan2023language}, CSG \cite{xinxin2020csg}, and TrafficGen \cite{feng2023trafficgen}, focus on generating specific traffic scenarios but often rely on real-world data or are limited by the functionalities of toolkits like ScenarioNet \cite{li2024scenarionet}. In contrast, AutoSceneGen bridges this gap by enabling users to efficiently collect AI-generated scenarios without relying on real-world data.

The concept of a \enquote{scenario} revolves around the \enquote{logic} governing objects within a scene. \citet{go2004elephant} defined a scenario as a detailed account involving actors, background details, objectives, and sequences of actions. In Autonomous Driving (AD), testing traditionally focuses on perception, motion planning, and decision-making modules \cite{huang2016avtest}. Simulators such as CARLA \cite{dosovitskiy2017carla} and AirSim \cite{shah2018airsim} have been developed to support these tests. \citet{goyal2023automatic} proposed a method to configure a scenario and automatically generate similar variants, while \citet{konig2024safedriving} introduced a formal model for test case generation and simulation verification. However, these approaches rely on abstract models, making them less scalable than prompt-based generation methods.

Several studies have explored augmenting traffic data and generating rare objects using ChatGPT \cite{alam2020auto, xinxin2020csg}. In this context, automatic scenario generation refers to creating scenarios from user prompts without detailed manual intervention. \citet{zhang2023find} reviewed the absence of automatic pipelines for safety-critical scenario generation in AD, highlighting challenges such as fidelity, efficiency, diversity, controllability, and transferability \cite{ding2022survey}. For example, \citet{yu2020multi} and \citet{sun2021corner} addressed challenges in generating safe lane-changing and adversarial cases, respectively.

\begin{table*}[!ht]
\centering
\setlength{\tabcolsep}{1.5mm}
\begin{tabular}{ccccccccccc}
\hline
Count          & UCY & ETH  & NGSIM  & KITTI & ApolloScapes & WaymoOpen & nuScenes & Argoverse2 & AutoSceneGen \\
\hline
duration (min) & 28  & 25   & 45     & 22    & 155          & 34200       & 0.34  & 458.34    & 60   \\ 
frames (×10\textsuperscript{3}) & 3000  & 3750  & 11.2  & 13.1  & 93.0   & 20k & 40 & 27500   & 0.05  \\ 
\hline
\multicolumn{10}{c}{total (×10\textsuperscript{3})} \\ 
\hline
pedestrian    & 120  & 0.65 & 0      & 0.09  & 16.2  & N/A & 222.164  & 20   & 0.25     \\ 
bicycle       & 0    & 0    & 0      & 0.04  & 5.5   & N/A & 11.859   & 12.5 & 0.05     \\ 
vehicle       & 0    & 0    & 2.91   & 0.93  & 60.1  & N/A & 1166.187 & 10k & 0.1      \\ 
\hline
\multicolumn{10}{c}{average (1/f)} \\ 
\hline
pedestrian    & 0.04 & 0.18  & 0      & 1.3   & 1.6  & N/A & N/A & N/A & 250      \\ 
bicycle       & 0    & 0     & 0      & 0.24  & 1.9  & N/A & N/A & N/A & 50       \\ 
vehicle       & 0    & 0     & 845    & 3.4   & 12.9 & N/A & N/A & N/A & 100      \\ 
\hline
\multicolumn{10}{c}{device} \\ 
\hline
Camera        &$\checkmark$&$\checkmark$& $\checkmark$ & $\checkmark$ & $\checkmark$ & $\checkmark$  & $\checkmark$ & $\checkmark$ & $\checkmark$   \\ 
LiDAR         & $\times$   & $\times$   & $\times$     & $\checkmark$ & $\checkmark$ & $\checkmark$  & $\checkmark$ & $\checkmark$ & $\checkmark$   \\ 
GPS           & $\times$   & $\times$   & $\times$     & $\checkmark$ & $\checkmark$ & $\times$      & $\checkmark$ & $\times$     & $\checkmark$   \\ 
Radar         & $\times$   & $\times$   & $\times$     & $\times$     & $\checkmark$ & $\times$      & $\checkmark$ & $\times$     & $\checkmark$   \\ 
IMU           & $\times$   & $\times$   & $\times$     & $\times$     & $\times$     & $\times$      & $\checkmark$ & $\times$     & $\checkmark$   \\ 
GNSS          & $\times$   & $\times$   & $\times$     & $\times$     & $\checkmark$ & $\times$      & $\times$     & $\times$     & $\checkmark$   \\ 
Lane          & $\times$   & $\times$   & $\times$     & $\times$     & $\checkmark$ & $\checkmark$  & $\times$     & $\checkmark$ & $\checkmark$   \\ 
Collision     & $\times$   & $\times$   & $\times$     & $\checkmark$ & $\times$     & $\times$      & $\times$     & $\times$     & $\checkmark$   \\ 
\hline
\end{tabular}
\caption{
A comparison of acquisition time, total frames, total instances, average instances per frame, and acquisition devices for the UCY, ETH, NGSIM, KITTI \cite{geiger2013visionkitti} (with trajectories), ApolloScapes datasets, and the AutoSceneGen's. Unlike other datasets with static volume data, AutoSceneGen's values are dynamically calculated based on speed per minute. The data reflects static speed during collection and includes trajectories with 8 attributes. Data collection was performed on a machine with an NVIDIA GeForce RTX 3090 GPU, 12th Gen Intel(R) Core(TM) i7-12700, using CARLA version 0.9.13 with the \enquote{Town03} map ($300 \times 300$).
}
\label{tab:dataset_comparisons1}
\end{table*}

Table~\ref{tab:dataset_comparisons1} compares the dataset generated by the AutoSceneGen with benchmark datasets, focusing primarily on AutoSceneGen's speed of data collection and diversity. The comparison emphasizes the framework’s efficiency in data collection (measured per minute) and its ability to capture diverse data forms. This highlights AutoSceneGen’s scalability, flexibility, and effectiveness in generating varied scenario types at scale. Unlike manually collected datasets, AutoSceneGen demonstrates superior adaptability and speed in producing diverse scenarios dynamically.
Though our framework supports collecting various sensor data, including LiDAR, this paper focuses on collecting trajectory data to demonstrate its capabilities.


\subsection{In-Context Learning}  
Do LLMs learn new tasks during inference, or do they simply recognize patterns from training? \citet{mann2020language} proposed that ChatGPT-3 can acquire new tasks through ICL, where task examples are embedded in prompts. They questioned whether models genuinely learn or merely recall familiar patterns, noting that synthetic tasks like word scrambling might be learned anew. However, doubts persist about whether LLMs truly understand prompts. For instance, \citet{webson2021prompt} found that models performed similarly with irrelevant prompts, suggesting that improved performance might not equate to genuine comprehension.
\citet{ming2022rethinkicl} argued that ICL depends more on input-label mapping, text distribution, and format than on actual ground-truth demonstrations, implying that ICL is ineffective without pre-existing input-label relationships. They also showed that ground-truth outputs may not be essential for tasks like generation. Their findings on NLP benchmarks leave unresolved questions about ICL's effectiveness in open-set tasks, its learning capacity during inference, and challenges like input-label mapping extraction. Additionally, \citet{dai2022whygpt} suggested that ChatGPT-3's ICL might function through implicit gradient descent as a meta-optimizer.

\textbf{The core basis of the AutoSceneGen} leverages the ICL capability of LLMs to generate optimized scenario configurations from given templates, simulating these scenarios in batch within a physics-based simulator. The generated scenarios reflect realistic physical interactions, ensuring their relevance for AV safety evaluation. To our knowledge, there is a lack of a universal framework that generates abundant, safety-critical, and realistic traffic scenarios specifically tailored for the safety evaluation of AVs.

\subsection{Rare Objects in An Open World}
\begin{figure*}[ht]
    \centering
    \begin{subfigure}[b]{0.245\textwidth}
        \centering
        \includegraphics[scale=0.0913, keepaspectratio]{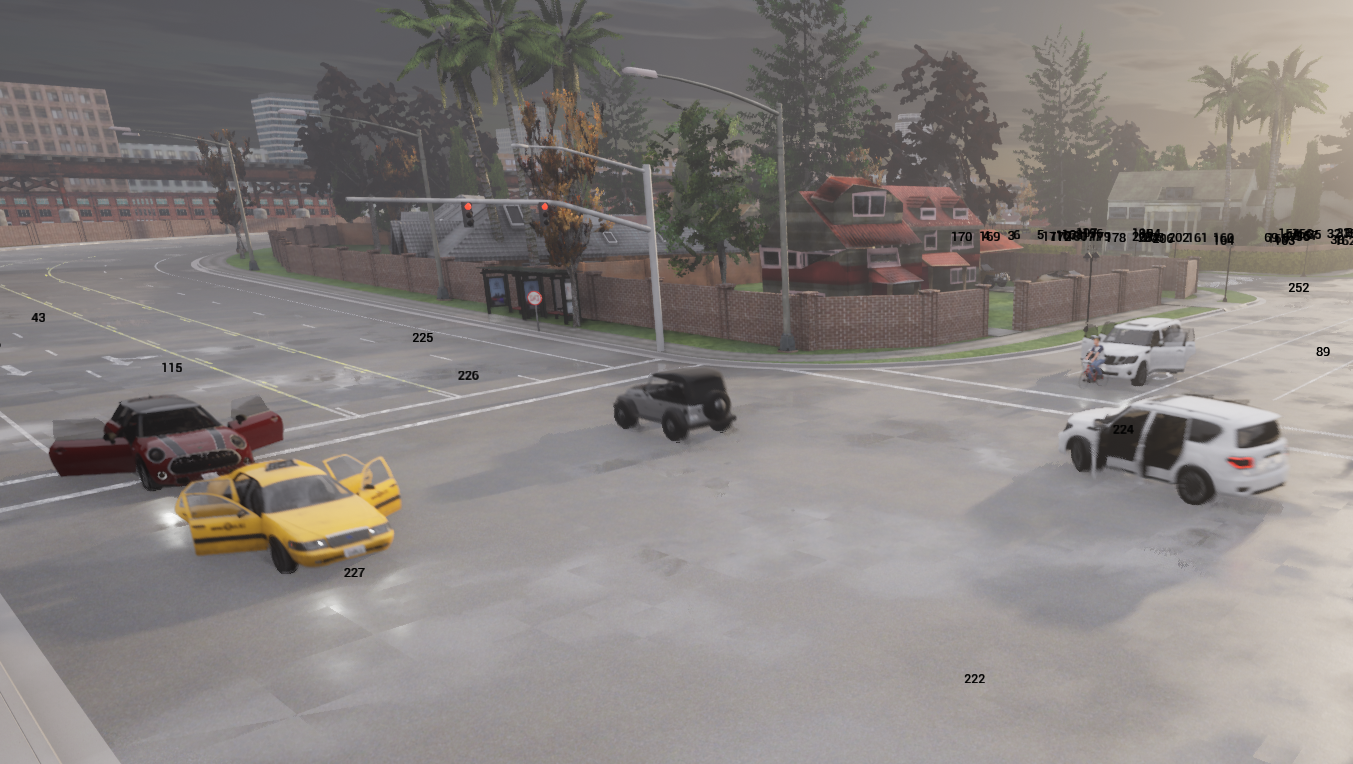}
    \end{subfigure}
    \begin{subfigure}[b]{0.245\textwidth}
        \centering
        \includegraphics[scale=0.0913, keepaspectratio]{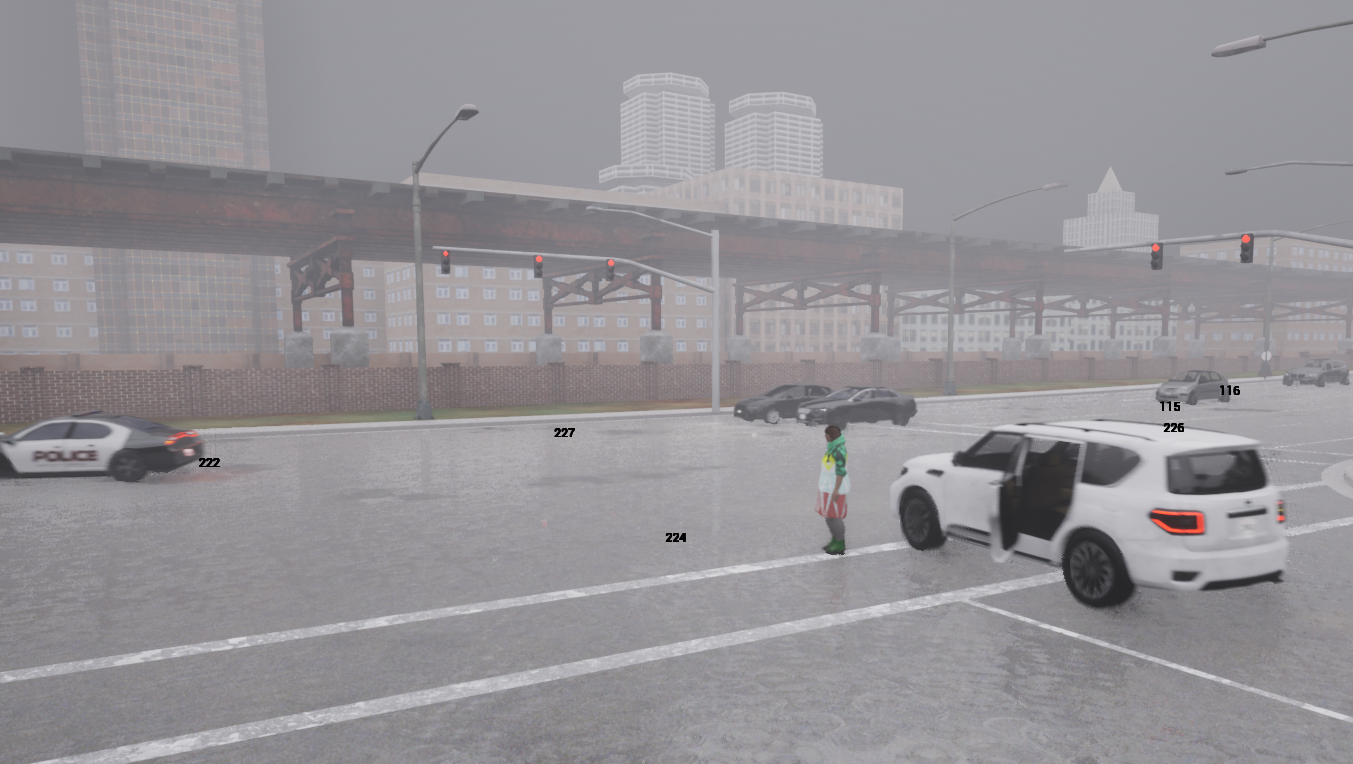}
    \end{subfigure}
    \begin{subfigure}[b]{0.245\textwidth}
        \centering
        \includegraphics[scale=0.0913, keepaspectratio]{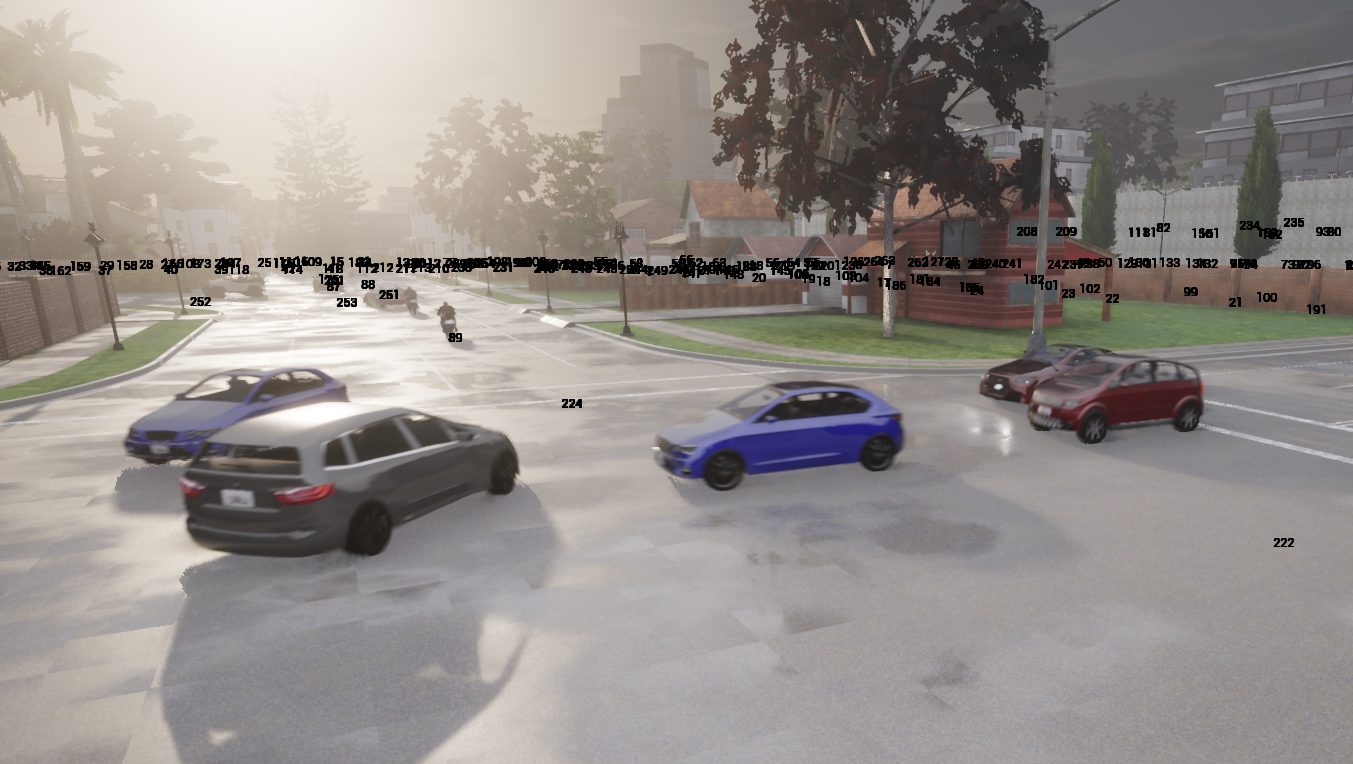}
    \end{subfigure}
    \begin{subfigure}[b]{0.245\textwidth}
        \centering
        \includegraphics[scale=0.0913, keepaspectratio]{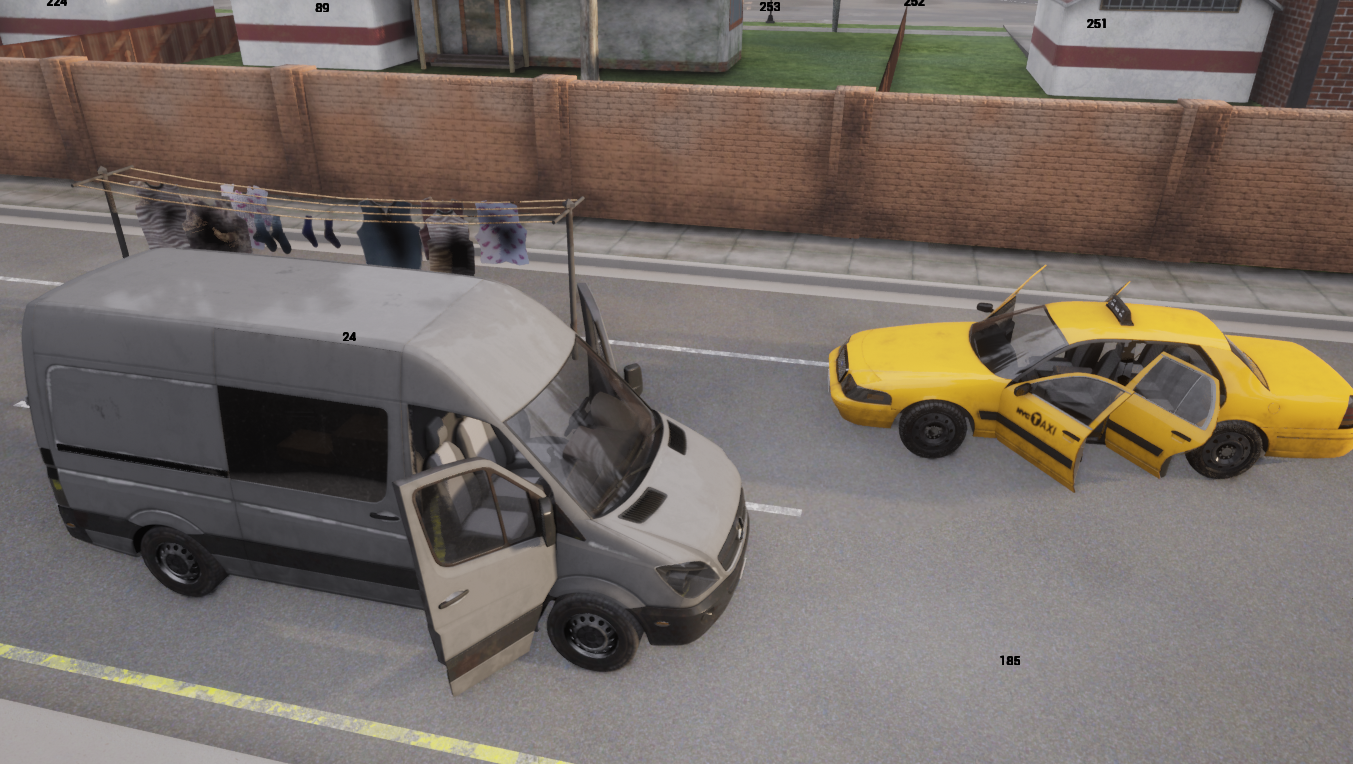}
    \end{subfigure}
    \caption{
    Images captured at four distinct timestamps and locations, corresponding to the AutoSceneGen input scenario description: \enquote{In downtown area, during a drizzly noon, there are vehicles malfunctioning windshield wipers and some of the vehicles' doors are open. Some vehicles exhibit negligent driving behavior, compromising visibility in wet conditions. There are 10 pedestrians on the road, with 50\% of the pedestrian running. No one was hurt and no accident happened since all the vehicles except the malfunctioning one obeyed the traffic rules.}
    }
    \label{fig:rare_scene}
\end{figure*}

Rare class objects, often overlooked in open-world data, are critical for ensuring model robustness in motion planning and visual detection. Manually creating such scenarios requires significant human effort, including writing detailed configurations, formulating scenario logic, and ensuring compatibility with the chosen simulator. These tasks demand extensive brainstorming and collaboration, particularly for rare and complex scenarios. For example, while an engineer might easily conceive a scenario involving a wrongly-parked vehicle with its doors open, envisioning affiliated events—such as a driver exiting the vehicle to chase someone or experiencing a sudden medical emergency—requires additional effort and time for ideation and planning. Such edge cases, though rare, are essential for the safety evaluation of AVs. 
This study addresses the challenge by leveraging LLMs' ICL capabilities to generate tailored configurations for rare scenarios, streamlining the ideation and scenario creation processes. Figure~\ref{fig:rare_scene} shows a rare scenario generated with this approach.

\subsection{Scenario Generation Efficiency}

Advancements in scenario generation have primarily focused on enhancing domain-specific languages such as Scenic \cite{fremont2019scenic} and ASAM OpenScenario. However, leveraging LLMs for scenario generation remains relatively unexplored. \citet{li2024chatgpt} introduced a framework utilizing ChatGPT to generate trajectory data, but its raw outputs lack physical realism and consistency with real-world scenarios, requiring active user intervention through Chain-of-Thought prompting \cite{zhang2022automatic}.
TARGET \cite{deng2023target} have utilized LLMs for generating simulator scenarios by incorporating a domain-specific language to create configurations based on self-defined traffic rules. However, their approach relies on a newly introduced XML-based domain-specific language, requires continuous manual input and interpretation, and has not been systematically evaluated for efficiency or scalability. Similarly, other methods, such as abstract-to-scenario and concrete-to-scenario approaches proposed by \citet{majzik2019sm}, face limitations in automating scenario creation for accuracy-critical domains like AD. 
Our framework offers a universal, cost-effective solution to enhance traffic scenario heterogeneity and facilitate the generation of rare, safety-critical scenarios in open-world environments, thereby streamlining the simulation and testing processes.


\begin{figure*}[!ht]
    \centering
    \includegraphics[scale=0.273]{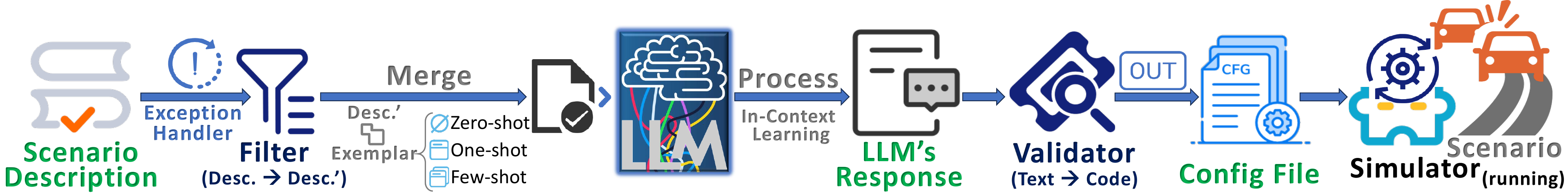}
    \caption{
    Architecture Overview. It begins with the user inputting a scenario description, which is managed by the \textit{Exception Handler} to block adversarial or irrelevant inputs, ensuring the framework operates within scope and prevents downstream issues. The \textit{Filter} processes the description, replacing simulator-incompatible terms with those aligned to the simulator's documented APIs. The filtered description (\textit{Desc.'}) is combined with pre-constructed ICL exemplars, which can be zero-shot, one-shot, or few-shot in category, depending on the LLM's familiarity with the simulator's APIs and the complexity of the scenario. The LLM generates a response containing scenario configurations, often accompanied by explanations and comments. The \textit{Validator} verifies each API call for compatibility, replacing unsupported terms with suitable alternatives (e.g., replacing \enquote{storm,} unsupported in CARLA, with \enquote{rain}) or ignoring them to prevent errors. This ensures all calls align with the simulator's capabilities, enabling execution of the final configuration file. The simulator runs the scenario, with the final step depicting the interaction between the real world and the virtual environment, while data collection can take place either inside the simulator (as is the case in this study) or externally.
    }
    \label{fig:aigs_architecture}
\end{figure*}

\section{In-Context Scenario Generation}
Existing efforts in driving scenario generation have primarily focused on creating videos to assess the perceptual capabilities of AVs. Recent advancements in incorporating rare object synthesis into realistic images for driving scenario generation have shown promise, offering valuable insights for improving downstream tasks. However, the exploration of AI-driven content creation has broadened our perspective, moving beyond the traditional concept of a \enquote{video} as a scenario. The core of a \enquote{scenario} lies in the underlying logic that governs the existence and interaction of objects within it. Reflecting on past approaches, we sought a more universal method to define the logic behind object interactions for scenario generation, rather than focusing solely on the static objects themselves. This distinction sets our work apart from the scope of AI-generated content (AIGC), which typically involves optimizing workflows or synthesizing objects into images for video creation aimed at training sensors and detectors. However, AIGC requires the training of new models, a common challenge across various domains, highlighting the need for models tailored specifically to the automotive industry to ensure accuracy and control over generated content.

While training a comprehensive model may provide a reliable strategy for generating accurate, domain-specific content, sourcing the data necessary for model training can be challenging. Additionally, motion planners trained on datasets directly collected from the real world may struggle to generalize effectively to unseen tasks, impacting model robustness, especially given the long-tail distribution of open-world cases. Moreover, training models from scratch demands significant time and computational resources, presenting challenges in the context of AD. Therefore, this research aims to explore alternative scenario generation methods that bypass the need for training or fine-tuning AI models, offering a more efficient and scalable approach to generating driving scenarios.

With ICL, the need for continuous fine-tuning or training of separate models for generation tasks is largely eliminated. Instead, LLMs equipped with ICL capabilities can be leveraged effectively. These models extend beyond traditional applications, such as language translation, to encompass more novel tasks such as code generation and scenario generation. This section focuses on the methodologies employed by AutoSceneGen, which effectively integrates ICL within the domain of LLMs. The integration of LLMs is anticipated to enhance both the efficiency and controllability of AutoSceneGen, offering a more streamlined and adaptable approach to scenario generation.

\subsection{System Architecture}

As shown in Figure~\ref{fig:aigs_architecture}, the system architecture consists of key components for processing scenario descriptions, which can be provided by the user or extracted from images using a vision-language model. A filtering process ensures simulator compatibility by replacing incompatible terms with appropriate alternatives. For example, in the description \enquote{in stormy weather,} the term \enquote{stormy} is identified as incompatible and replaced with \enquote{rainy.} This replacement is performed using a predefined replacement dictionary derived from the simulator documentation, as detailed in Algorithm~\ref{alg:filter}.

\begin{algorithm}[ht]
\caption{Filter for Scenario Descriptions}\label{alg:filter}
\KwIn{Raw Scenario Description $D$}
\KwOut{Final Scenario Description $D'$}
\For{each token $t_i \in D$}{
    \eIf{$t_i \in \text{SimulatorDocument}$}{
        $t_i \gets \text{SimulatorDocument}(t_i)$\;
    }{
        \If{$t_i \in \text{Words}^{\text{incorrect}} \cup \text{Punctuations}^{\text{wrong}}$}{
            Correct $t_i$\;
        }
    }
}
\end{algorithm}


The user’s scenario description is combined with few-shot learning examples within the framework. Developers adapting the framework to their specific domain need to create few-shot learning exemplars based on simulation needs. The process of constructing these exemplars depends on the chosen LLM: if all relevant APIs and examples are included in the LLM's train set, no additional exemplars are required. However, if the necessary documentation is not part of the train set, one-shot or few-shot exemplars need to be pre-constructed to meet the simulation requirements. The LLM then generates the configuration by leveraging both the scenario description and the ICL examples to ensure an accurate simulation.

\begin{algorithm}[t]
\caption{Validator for Generated Configuration}\label{alg:validator}
\KwIn{Generated configuration $C_{LLM}$}
\KwOut{Validated code $F$}
Extract code from $C_{LLM}$ to get $F$\;
\For{each term $t_i \in C_{LLM}$}{
    \If{$t_i \notin \text{SimulatorDocument}$}{
        Search for synonym $t_i^{syn}$\;
        \eIf{$\exists t_i^{syn} \in \text{SimulatorDocument}$}{
            Replace $t_i$ with $t_i^{syn}$\;
        }{
            Remove $t_i$\;
        }
    }
}
$F \gets C_{LLM} \setminus \{\text{comments}, \text{explanations}\}$\;
\end{algorithm}


Prior to execution, the validator (as detailed in Algorithm~\ref{alg:validator}) checks each API call against a documented list, replacing unsupported terms with approved synonyms (e.g., replacing \enquote{Weather.Storm} with \enquote{Weather.Rain} in CARLA). If no replacement is found, the term is ignored to prevent errors, ensuring the calls align with the simulator's capabilities. The validator also ensures that the final configuration includes only relevant code, removing any extraneous content such as explanations and comments, and verifies adherence to the syntax required by the simulator.  

Finally, the simulator is launched to run the generated configuration files, ensuring seamless execution within the simulator itself or on connected digital twin platforms. Our framework automates AV evaluation by efficiently generating traffic scenarios from user-defined descriptions, ensuring cost-effective simulation across diverse open-world environments.

\textbf{Input.} The user provides scenario descriptions.

\textbf{Process.} The scenario descriptions are combined with simulator-specific examples and preprocessed to ensure clarity and compatibility. The chosen LLM then processes the input. The validator checks the generated configuration for adherence to the simulator's syntax and grammar, ensuring flawless execution.

\textbf{Output.} A scenario being simulated, provided by an executable configuration file, which is used by the simulator in either its purely virtual environment or on a digital twin platform connected to it.

\subsection{LLMs \text{\&} ICL}
Since LLMs exhibit ICL, they can handle tasks they haven't been explicitly trained on without the need for retraining or fine-tuning, allowing them to perform well on specialized tasks using a few-shot learning approach.

{\bf Input.} In the scenario generation process, the input consists of prompts (scenario descriptions). Additionally, to adapt the model to a specialized task, ICL examples that match the desired range of outputs must be provided alongside the scenario description as input to the LLMs.

{\bf Output.} The response may include essential scenario elements directly usable by the simulator, as well as explanations that cannot be executed. To refine the response, two approaches are considered: manually scripting to isolate executable code or using the LLM to automatically extract only the code. It is generally believed that with proper ICL examples, the LLM will output only executable code without additional information. However, in the experiment, we found that even when we accentuated the requirement of \enquote{do not generate any explanations or comments except code} to the chosen LLMs, the output may still contain phrases like \enquote{Here is the result....} Additionally, considering that the code generated by LLMs undergoes grammar validation in the subsequent process, useful content extraction from its responses can also be accomplished through a manually scripted validation procedure. This approach ensures that the input prompt remains streamlined while preserving tokens for efficient processing.


\begin{table*}[!ht]
\centering
\setlength{\tabcolsep}{3mm}
\begin{tabular}{lccccccccc}
\toprule
Dataset                 & Method         & TAE   & ADEv    & ADEp   & ADEb   & TFE    & FDEv  & FDEp & FDEb    \\
\midrule
A.S.                    & TrafficPredict & 0.085  & 0.080  & 0.091  & 0.083  & 0.141  & 0.131 & 0.150 & 0.139  \\
A.S. train-set + Ours   & TrafficPredict & 0.053  & 0.085  & 0.058  & 0.065  & 0.076  & 0.114 & 0.092 & 0.094  \\   
Ours                    & TrafficPredict & 0.033  & 0.088  & 0.020  & 0.047  & 0.058  & 0.135 & 0.037 & 0.077  \\   
\midrule
TRAF                    & TraPHic        & 5.63   &N/A     &N/A     &N/A     & 9.91   &N/A     &N/A     &N/A      \\
A.S.(Reproduced)        & TraPHic        & 5.10   & 3.62   & 1.02   & 4.49   & 2.81   & 6.73   & 1.88   & 8.44    \\
A.S. train-set + Ours   & TraPHic        & 1.30   & 1.69   & 0.42   & 0.90   & 2.10   & 2.82   & 0.67   & 1.39    \\
Ours                    & TraPHic        & 0.27   & 0.14   & 0.19   & 0.44   & 0.40   & 0.21   & 0.30   & 0.62    \\
\bottomrule
\end{tabular}
\caption{
The comparison results of the ApolloScapes dataset, collected from AutoSceneGen (Ours), and the two datasets combined are presented. The term \enquote{ApolloScapes Dataset} is abbreviated as \enquote{A.S.} in the table. Lower metrics indicate better performance. We generated 17,919 examples; the official training set of A.S. contains 94 examples. We used the original method proposed in TrafficPredict to train the planner. For TRAF and another evaluation, we used TraPHic. While the results are not as good as the results under TrafficPredict, under the method TraPHic there are still huge improvements thanks to the substitution of TRAF \cite{chandra2019traphic} and ApolloScapes \cite{ma2019trafficpredict} to the dataset collected via AutoSceneGen. Suffix \enquote{v}, \enquote{b} and \enquote{p} in \enquote{ADE}/\enquote{FDE} represents \enquote{vehicle}, \enquote{bicycle} and \enquote{pedestrian} respectively.
}
\label{tab:dataset_comparisons2}
\end{table*}

\section{Evaluations}
The datasets generated using the AutoSceneGen can be extensive, with its simulator-dependency offering significant flexibility and scalability, which allows data to be collected for a wide range of scenarios as long as the simulator supports the required functions. 
By running the scenarios generated by the framework, diverse datasets can be efficiently created for various purposes and different applications. 
To evaluate AutoSceneGen, we utilized existing approaches for trajectory prediction. 
In the first step, we replaced their real datasets with our own; in the second step, we combined their datasets with ours. Finally, we tested exclusively with our own data. This allowed us to determine whether our data could achieve results comparable to those obtained with the original datasets. As an example, we used AutoSceneGen with GPT-4 \cite{achiam2023gpt4} as the selected LLM to generate 125 traffic scenarios, each based on a short prompt that described the scenario we aimed to generate.
To evaluate how our work can expedite the safety evaluation process for AVs, we selected a subset of safe scenarios from the 125 scenario configurations, then loaded them into the simulator to simulate the generated scenarios. 
Next, we used the data collector to gather data in the required format and style, based on the comparison results that we would later use for evaluation. By directly replacing only the train set of the target dataset, we initiated the evaluation process of the scenarios collected through the AutoSceneGen framework using existing trajectory prediction approaches.
Finally, after collecting the data in the required format and training the model from scratch to observe the trajectory prediction results, we directly added our dataset, collected through the AutoSceneGen, to the original training set used primarily for the evaluation of trajectory prediction methods by their proposers.

Not only can AutoSceneGen achieve the massive and efficient collection of critical scenarios, but the datasets it generates are also logically realistic and heterogeneous.
In the experiment, we ran 41 AI-generated scenarios with varying numbers of vehicles and pedestrians to collect as much data as possible under different circumstances within a single scenario logic, using the official CARLA map from Town01 to Town05.

We evaluated our dataset using average displacement error (ADE) and final displacement error (FDE). We replicated TrafficPredict's results \cite{ma2019trafficpredict} with their ApolloScapes dataset, which features curated trajectory data from real road scenarios. For TRAF \cite{chandra2019traphic}, lacking access to their full dataset, we used ApolloScapes with the \enquote{TraPHic} method. This approach allowed us to assess the dataset collected from AutoSceneGen across different methods.

\begin{figure*}[!ht]
    \centering
    \begin{subfigure}[b]{0.23\textwidth}
        \centering
        \includegraphics[scale=0.21, keepaspectratio]{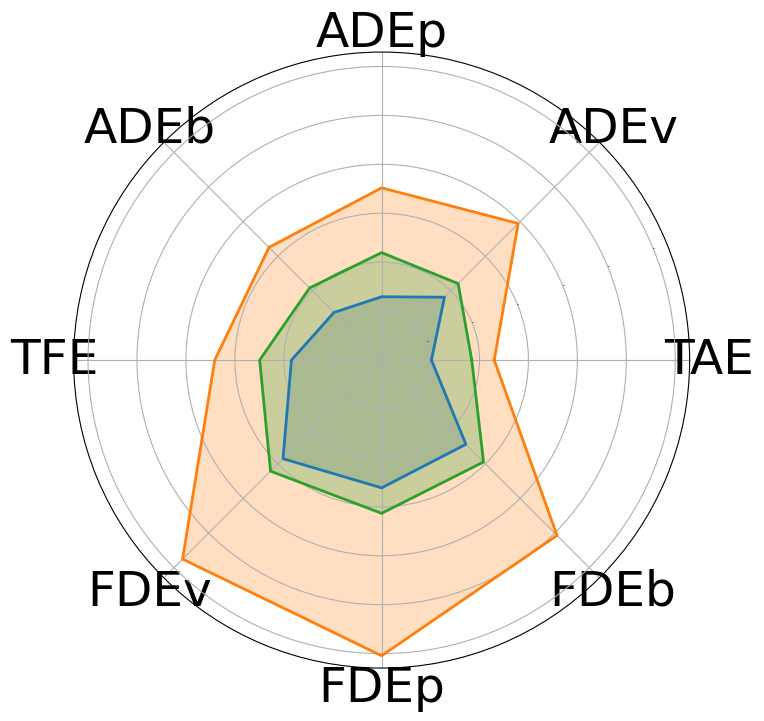}
        \caption{
        Epoch 127}
        \label{fig:abc_epoch_127}
    \end{subfigure}
    \begin{subfigure}[b]{0.23\textwidth}
        \centering
        \includegraphics[scale=0.21, keepaspectratio]{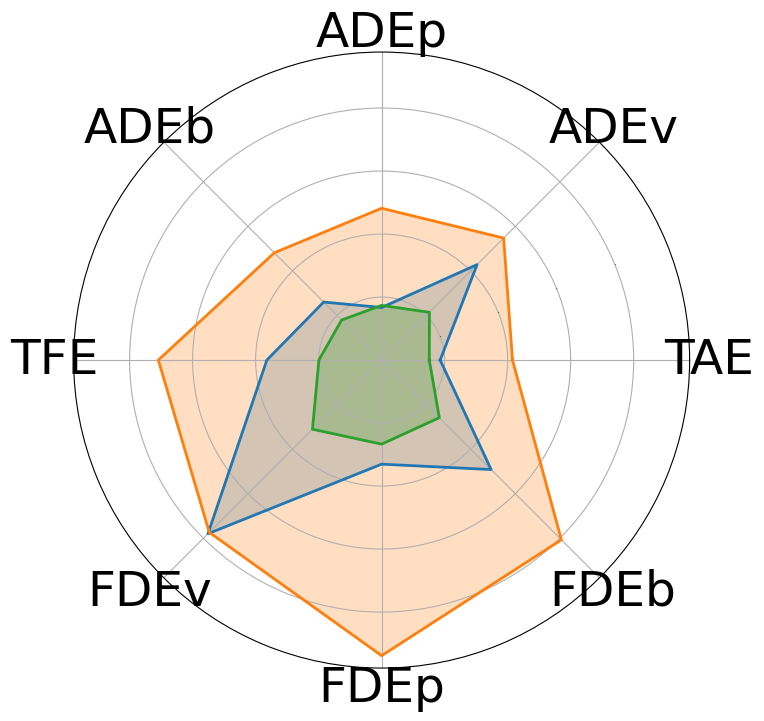}
        \caption{
        Epoch 147}
        \label{fig:abc_epoch_147}
    \end{subfigure}
    \begin{subfigure}[b]{0.23\textwidth}
        \centering
        \includegraphics[scale=0.21, keepaspectratio]{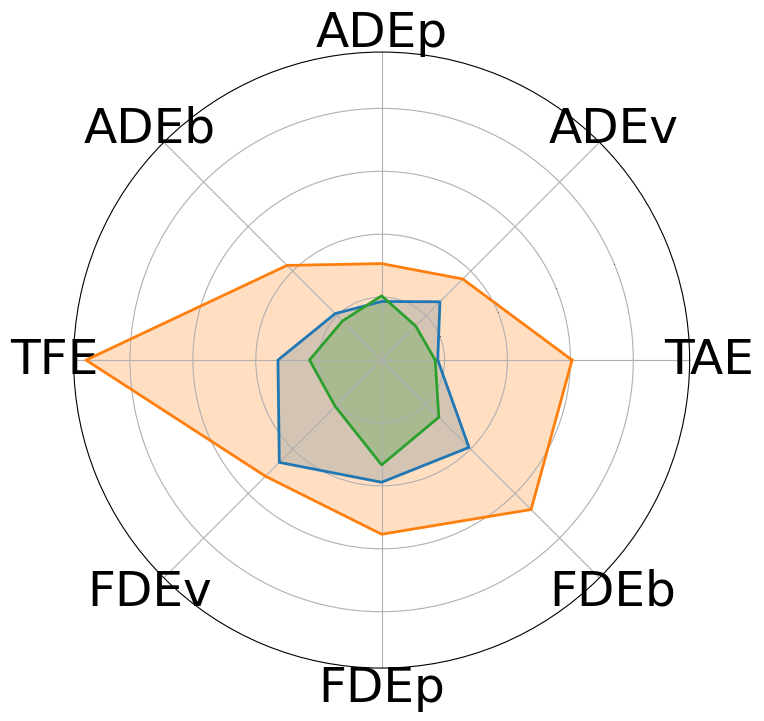}
        \caption{
        Epoch 160}
        \label{fig:abc_epoch_160}
    \end{subfigure}
    \begin{subfigure}[b]{0.23\textwidth}
        \centering
        \includegraphics[scale=0.21, keepaspectratio]{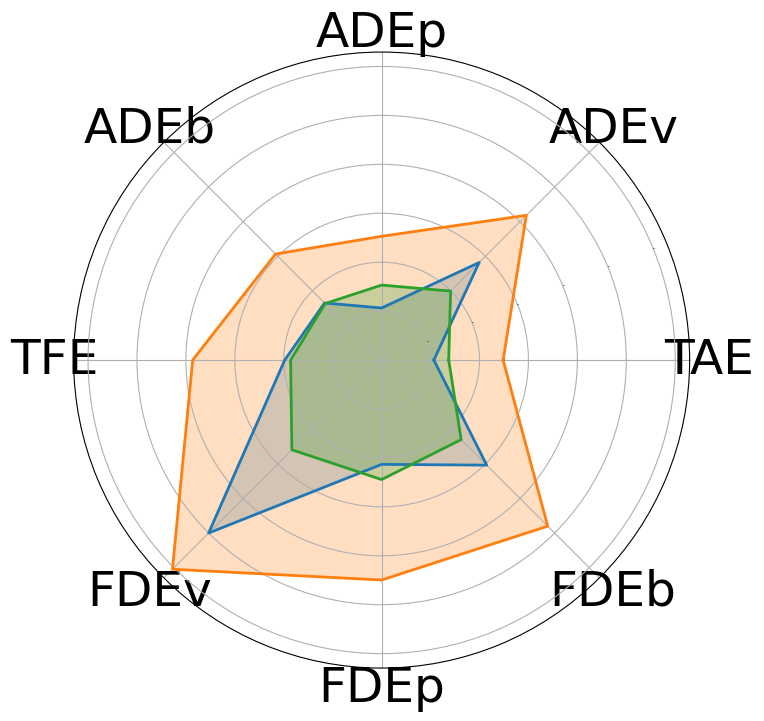}
        \caption{
        Epoch 200}
        \label{fig:abc_epoch_200}
    \end{subfigure}
    \caption{
    The comparison of all metrics between the datasets collected via AutoSceneGen (Blue), ApolloScapes (Orange), and the combination of the two datasets (Green) across different epochs is shown. While the dataset collected purely from AutoSceneGen outperforms ApolloScapes in some epochs, such as epoch 127, the combination of AutoSceneGen and ApolloScapes demonstrates better overall results. Due to the distinct distribution of traffic participants in the two datasets, Figures (b), (c), and (d) show sharper peaks for FDE-vehicle and ADE-vehicle. However, the combination of the two datasets achieves reasonable values overall. In this experiment, ApolloScapes has a total of 3,917 frames, AutoSceneGen has 17,919 frames, and the combined AutoSceneGen + ApolloScapes has 27,605 frames.
    }
    \label{fig:abc_comparison2}
\end{figure*}

Without modifying the original trajectory prediction network, our dataset achieved superior results with reduced displacement error for each traffic participant type, as shown in Table~\ref{tab:dataset_comparisons2}. In various epochs, the dataset collected from AutoSceneGen demonstrated the highest accuracy in trajectory prediction, as illustrated in Figure~\ref{fig:abc_comparison2}-(a). Moreover, combining our dataset with ApolloScapes improved overall performance, enhancing all trajectory prediction metrics by incorporating diverse scenarios and extensive data, as depicted in Figures~\ref{fig:abc_comparison2}(b), (c), and (d).

Besides ApolloScapes, several experiments were also conducted on TraPHic, Pishgu \cite{alinezhad2023pishgu}, and MSRL \cite{wu2023multi} using other datasets in a similar manner. Since the NGSIM dataset contains only vehicle trajectories and the ETH and UCY datasets contain only human trajectories, we excluded extra calculations on unseen participants in each comparison. For VIRAT \cite{oh2011large} and ActEV \cite{awad2020trecvid}, although they include three major event types (including vehicles), we focused only on the annotated pedestrian trajectories to control for variables when comparing under Pishgu. Table~\ref{tab:dataset_comparisons3} presents a comparison of the results between our dataset and NGSIM, ETH/UCY, and VIRAT/ActEV. Our dataset, collected via AutoSceneGen using CARLA 0.9.13, includes 264 pedestrian trajectories (1,848k frames) and 770 vehicle trajectories (15,400k frames), with the vehicle trajectories replacing and augmenting NGSIM data.

\begin{table}[!ht]
\centering
\begin{tabular}{lccc}
\hline
Dataset                 & Method         & ADE    & FDE   \\
\hline
NGSIM                   & Pihgu          & 0.88   & 1.96   \\  
Ours                    & Pihgu          & 7.98   & 15.43  \\
NGSIM train-set + Ours  & Pihgu          & 0.84   & 1.87   \\
ETH/UCY                 & Pihgu          & 1.10   & 2.24   \\
Ours                    & Pihgu          & 1.48   & 2.70   \\ 
ETH/UCY train-set + Ours& Pihgu          & 0.79   & 1.50   \\
VIRAT/ActEV             & Pihgu          & 14.11  & 27.96  \\ 
Ours                    & Pihgu          & 16.05  & 31.09  \\
VIRAT/ActEV + Ours      & Pihgu          & 15.32  & 29.65  \\ 
\hline
\end{tabular}
\caption{
The comparison results of the NGSIM dataset, the dataset collected from AutoSceneGen (ours), and the combination of the two datasets are shown. When replacing the NGSIM dataset with ours, the ADE and FDE values are much higher (worse) than when using the original NGSIM dataset. However, when we combine the two datasets—NGSIM and ours—the ADE and FDE decrease and outperform the results obtained from using the NGSIM dataset alone, indicating that the original NGSIM dataset is augmented by our dataset collected via AutoSceneGen. A similar observation was made with the ETH/UCY dataset.
}
\label{tab:dataset_comparisons3}
\end{table}

Table~\ref{tab:dataset_comparisons3} demonstrates that the dataset generated by AutoSceneGen achieves comparable or even superior results on several state-of-the-art trajectory prediction methods when compared to those using real-world road datasets. While benchmark datasets in Table~\ref{tab:dataset_comparisons3} typically provide more diverse scenarios with manual annotations, the generated dataset is based on 125 scenario descriptions. This highlights the framework's effectiveness by revealing key trends in ADE and FDE. When combined with benchmark datasets, the generated data reduces displacement error by introducing valuable new scenarios. For instance, incorporating the generated data into NGSIM and ETH/UCY improves results compared to using either dataset alone, due to the complementary nature of the data rather than simply increasing volume. The generated dataset outperforms in most cases due to its scenario diversity. However, in VIRAT/ActEV, no significant improvement is observed, largely due to the chosen simulator's focus on vehicle simulation and limited pedestrian simulation capabilities. Nevertheless, the tendency for the generated data to outperform remains evident, suggesting potential for further improvement with more diverse pedestrian simulation.

\section{Limitations}
Our framework has several limitations, including performance, simulator integration, and data availability, as well as regulatory constraints due to varying global traffic laws. Developers may need to create custom virtual environments or choose different simulators if those mentioned in this paper do not meet their regulatory requirements. The framework's effectiveness depends on the capabilities of both LLMs and the simulator, particularly for tasks such as scenario generation and accident reconstruction. If the chosen LLM lacks ICL capabilities or the simulator cannot handle complex scenarios, the framework’s utility is significantly reduced, making it difficult to generate corner cases. Like many AIGC toolkits, the explainability of the generated scenario logic is limited to the explainability of the scenario descriptions and ICL examples, especially at higher abstraction levels. This limitation arises from the chosen LLMs, not from the framework itself. To address this, the framework enhances explainability through detailed configuration files stored after validation, allowing users to trace the rationale behind generated scenarios by comparing input prompts with generated configurations.

Last but not least, if the example data for ICL is insufficient, some simulators may not respond effectively, necessitating post-revision of the responses. Most example data is sourced from official documentation or the open-source community of the chosen simulator, but availability and usage rights are not always guaranteed.

\section{Conclusions}
AutoSceneGen framework is designed for the generating of traffic scenarios using foundation models. The resulting scenarios can be simulated within virtual environments in simulators, within the real world, or even both concurrently. The framework achieves the improvement of the work efficiency on generating sufficient safety-critical scenarios conveniently for testing AVs in an open world, as well as the convenience of traffic accident reconstruction.
As highlighted by \citet{kalra2016dirve}, AVs face the challenge of needing to cover vast distances — potentially hundreds of millions to even billions of miles — to establish their reliability in terms of minimizing fatalities and injuries. AutoSceneGen is particularly advantageous for enhancing the reliability testing of AVs slated for real-world deployment. It achieves this by harnessing the power of foundation models to simulate extensive arrays of traffic scenarios, thereby providing a robust testing environment for AVs.

\bibliography{aaai25}
    
\end{document}